\documentclass[conference]{IEEEtran}
\IEEEoverridecommandlockouts
\usepackage{cite}
\usepackage{amsmath,amssymb,amsfonts}
\usepackage{algorithmic, amssymb, gensymb}
\usepackage{graphicx}
\usepackage{textcomp}
\usepackage{xcolor}

\def\BibTeX{{\rm B\kern-.05em{\sc i\kern-.025em b}\kern-.08em
    T\kern-.1667em\lower.7ex\hbox{E}\kern-.125emX}}

\usepackage[caption=false]{subfig}

\usepackage[breaklinks=true,bookmarks=false]{hyperref}

\title{\LARGE \bf
ArrowPose: Segmentation, Detection, and 5 DoF Pose Estimation Network for Colorless Point Clouds
\thanks{This project was funded in part by Innovation Fund Denmark through the projects MADE ReAct and FERA, and in part by the SDU I4.0-Lab.}
}

\author{\IEEEauthorblockN{1\textsuperscript{st} Frederik Hagelskjær}
\IEEEauthorblockA{\textit{SDU Robotics, Mærsk Mc-Kinney Møller Institute} \\
\textit{University of Southern Denmark}\\
5230 Odense M, Denmark \\
frhag@mmmi.sdu.dk}
}

\begin{document}

\maketitle
\thispagestyle{empty}
\pagestyle{empty}

\begin{abstract}
This paper presents a fast detection and 5 DoF (Degrees of Freedom) pose estimation network for colorless point clouds. The pose estimation is calculated from center and top points of the object, predicted by the neural network. 
The network is trained on synthetic data, and tested on a benchmark dataset, where it demonstrates state-of-the-art performance and outperforms all colorless methods. The network is able to run inference in only 250 milliseconds making it usable in many scenarios. 
Project page with code at \url{arrowpose.github.io}
%
\end{abstract}
\begin{IEEEkeywords}
deep learning, pose estimation, point cloud
\end{IEEEkeywords}

\section{Introduction}


Pose estimation enables much greater flexibility in robotics. Objects can be placed freely in the environment and poses updated online. This is in contrast to using fixtures with manually inserted objects. Pose estimation allows the use of robots in situations where pre-configuration is either too expensive or impossible, enabling automation in a much larger realm of tasks. 
While historically a complicated task, pose estimation has been aided greatly by the introduction of deep learning \cite{hodan2018bop, sundermeyer2023bop, hodan2024bop}.
Object detectors are an integral part of modern pose estimation pipelines, as most pose estimation algorithms only work with cropped images of the relevant object \cite{hodan2024bop}.

However, object detectors for pose estimation generally use RGB information to process the scene \cite{hodan2024bop}. In current benchmarks of pose estimation algorithms \cite{hodan2018bop, sundermeyer2023bop, hodan2024bop} there are currently no depth-only object detectors. As a result, all depth-only methods are variants of the classic method Point-Pair-Feature (PPF) \cite{drost2010model}. Other depth-only pose estimation methods rely on RGB detectors in the pipeline \cite{redickbayesian, konig2020hybrid}.
However, for some use cases, color information is not available, either because of the sensors or the environment. Objects detectors for pose estimation using only depth is thus an important topic. While, detection algorithms for colorless point clouds have been explored extensively for processing scenes for autonomous cars and robots \cite{lang2019pointpillars, zhao2024autonomous, chen2023voxelnext}. These methods generally use lidar data and are not optimized for object detection for pose estimation. 





%

In this paper, we present a simple instance detector using only colorless point clouds as input. The network computes semantic segmentations and predicts the object centers. Instance segmentations are then computed by clustering the center predictions. 
The method is extended to 5 DoF pose estimation by also predicting the orientation vector of the symmetry axis. This pose estimation method only works for objects with cylindrical symmetry. However, many industrial objects fit this description; idlers, bearings, screws, etc. \cite{yokokohji2019assembly}, and such a method is thus relevant. Our method could be extended for 6 DoF pose estimation by predicting a orthogonal vector to the orientation vector. However, for many objects, this orientation cannot be estimated from the sensor data. Instead, attempting to estimate this aleatoric uncertainty, it has been removed from the model.  
The benefit of the 5 DoF pose estimation is demonstrated by the large increase in performance for the rotational symmetric object from the IC-BIN \cite{doumanoglou2016recovering} benchmark dataset. Our method also demonstrates state-of-the-art performance for the full dataset.


\begin{figure}
\centering
\subfloat[Point cloud.]{\includegraphics[width=.48\linewidth]{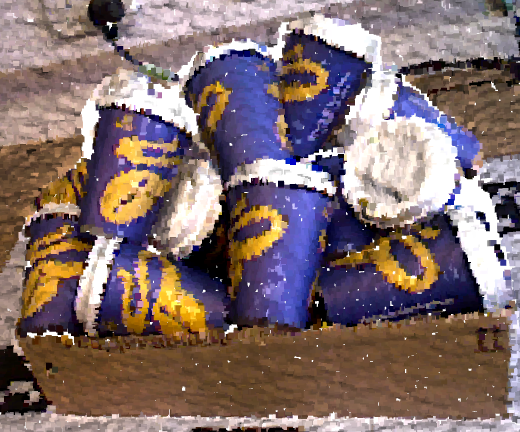}}\ 
\subfloat[Center and top predictions.]{\includegraphics[width=.48\linewidth]{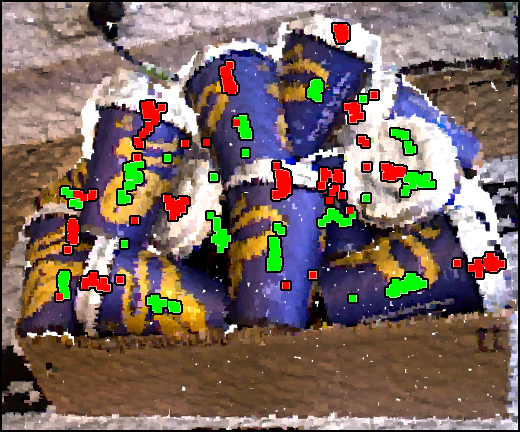}}

\subfloat[Center and orientation vector.]{\includegraphics[width=.48\linewidth]{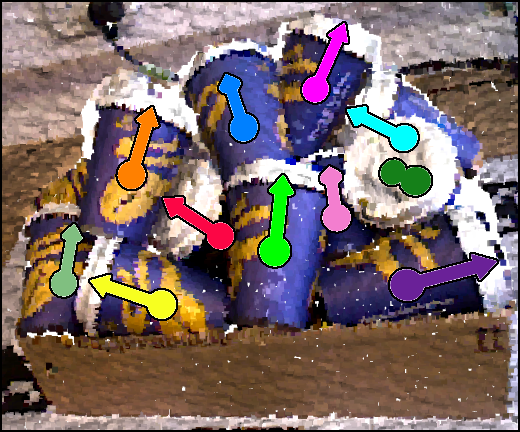}}\ 
\subfloat[Pose Estimation.]{\includegraphics[width=.48\linewidth]{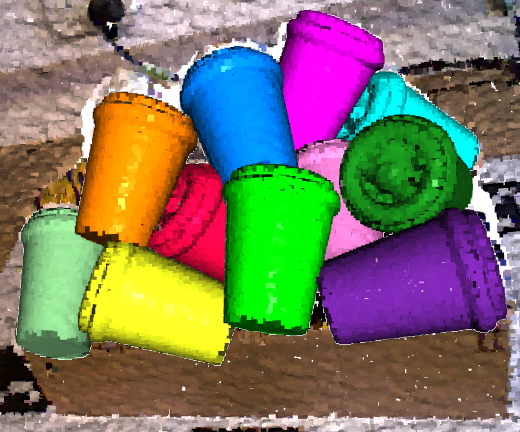}}

   \caption{The inference pipeline of our developed method. (a) Objects in the bin. (b) The neural network predicts the center and top points belonging to the object, shown in green and red. 
   (c) By clustering the predictions the objects 5 DoF poses are found as a center and an orientation vector.
   (d) Visualization of the pose estimate using the object model.
   %
   Colors only for visualization.}
   \label{fig:odpe}
     \vspace{-6mm}
\end{figure}

%
A method similar to our work is presented in \cite{li2025workpiece}. The input point cloud is processed by the PointPillars network \cite{lang2019pointpillars}, a network for processing large point clouds from lidar data. 
However the detections are not used for pose estimation, and instead the detections are used to compute grasp poses.
%
%

A method also performing object detection and pose estimation in point clouds is presented in \cite{zhuang2023instance}. The input to the network is a 16,384 point large point cloud which is preprocessed by a PointNet++ \cite{qi2017pointnet++} backbone. Our work, on the other hand, processes a larger point cloud of 65,536 points, and uses a different backbone. While both methods computes the semantic segmentations, the instance segmentation are computed differently. In our work, the instance segmentation is performed by clustering object centers, compared with a calculated feature distance in the paper. The paper calculates the object poses by two MLPs (Multi Layer Perceptron) that predicts the translation and rotation, whereas our method computes the center and orientation vector.
%
The method is compared with the PPF \cite{drost2010model} method on a synthetic dataset introduced in the paper. The method shows an average performance better than PPF \cite{drost2010model}, but scores lower for three of the four objects. In comparison our method is tested on a benchmark dataset with real data.

















The main contributions presented in this paper are:

\begin{itemize}
    \item A novel network for processing large point clouds with pre-processing on the CPU.
    \item A robust method for computing 5 DoF pose estimation from center  and top points. 
    \item State-of-the-art results on the IC-BIN \cite{doumanoglou2016recovering} benchmark dataset using depth only methods.
\end{itemize}


\section{Method}



This section provides a full description of our method. It starts with a description of the inference pipeline and then elaborates the training process.

\subsection{Inference pipeline}

\begin{figure}[]
    \vspace{1.5mm}
    \begin{center}
       \includegraphics[trim=0 0 0 0, clip,width=0.95\linewidth]{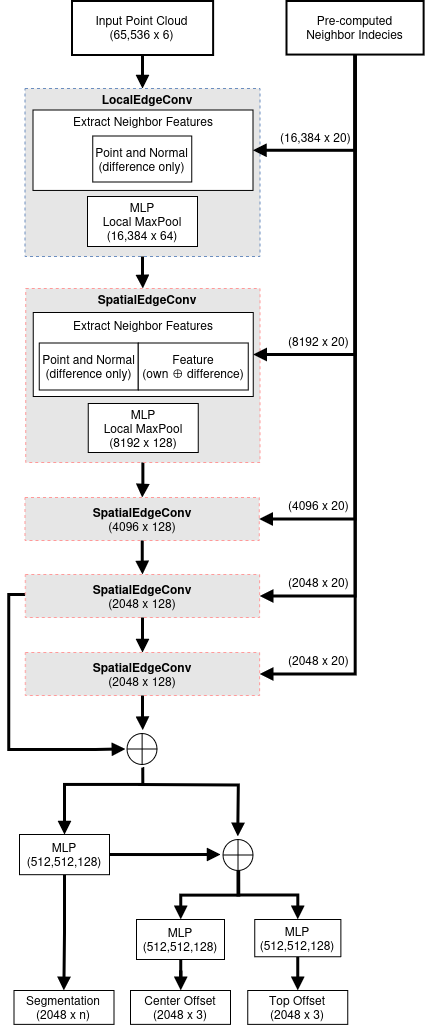}
       \caption{\textbf{Network Architecture:} The network input is a point cloud and pre-computed neighbor indices. For each layer neighbor differences are computed and processed, while the point cloud is downscaled. Finally, the features for the last two layers are concatenated, $\oplus$, and processed by an MLP. Using this, the segmentation prediction is computed for each $n$ class. The MLP output is concatenated with the earlier concatenated features and two MLPs predict the center- and top-offsets.
       }
       \label{fig:pyramid}
    \end{center}
\end{figure}

\subsubsection{Network structure}
The focus of the developed network is a small network for fast processing of large point clouds. The network is based on the structure of DGCNN \cite{dgcnn}, but uses the spatial pyramid of PointNet++ \cite{qi2017pointnet++}.
%
%

Similar to DGCNN \cite{dgcnn} we use the EdgeConv, which computes the feature difference with neighboring points. These differences are then concatenated with the points own feature and processed by a convolutional network, and finally by using a local max poll the information is reduced to a one dimensional feature. In the first layer of DGCNN the neighbors are found in the spatial domain, but in subsequent layers the neighbors are found in the feature domain to increase the receptive field information for each point. However, for very large point clouds this computation is unfeasible as it is performed with an adjacency matrix, which is $n \times n$, where $n$ is the number of points in the point cloud \cite{hagelskjaer2022deep}.

Instead we use two approaches from PointNet++ to allow us to process the large point clouds. Firstly, the neighbor points are computed in point distance instead of feature distance. This allows for pre-computing the neighbors on the CPU which is better suited for this task than the GPU as it does not require the adjacency matrix. Secondly, we obtain a larger neighbor distance for each point by down-sampling the point cloud during each EdgeConv computation. The down-sampling which is applied four times, resulting in a reduction of the point cloud from 65,536 to 2048 points. This also fits better with the segmentation task as points need information about the neighborhood, but global information is unnecessary. 
In DGCNN the EdgeConv computes the point information and the difference with its neighbors Eq.~(1), (Eq.~(7) in \cite{dgcnn}). Where $x_i$ is the local point and $x_j$ is a neighboring point. $\bar{h}_{\theta}$ is the trained convolution with a ReLU activation function followed by a MaxPool.
%
\begin{equation}
 \bar{h}_{\theta}(x_i,x_j-x_i) 
\end{equation}
%
However, for the first layer the features are the spatial position and normal vectors, meaning that $x_i$ encodes global position information. For our object detector this global information is unnecessary noise, as the detection is should not depend on global position. Instead we use Eq.~(2), (Eq.~(6) in \cite{dgcnn}) which only uses local information, in our model we denote this the LocalEdgeConv ($\bar{h}$ is shown as $h$ in \cite{dgcnn}).
\begin{equation}
\bar{h}_{\theta}(x_j-x_i)
\end{equation}
But, in the subsequent layers $x_i$ simply contains the feature computed for the point, and thus we want to retain this information. Additionally, as the global position information is recorded the relative position between points is unknown in the is kept. However, as the global positions of the points is not used the spatial relationship is not included in the features. To include this we add a new layer, here denoted the SpatialEdgeConv, as it computes the original EdgeConv, but also includes spatial difference about neighbors. The feature computation is shown in in Eq.~(3), where $p_i$ and $p_j$ indicates the spatial information for the point and neighbor respectively.
\begin{equation}
\bar{h}_{\theta}(p_j-p_i,x_i,x_j-x_i)
\end{equation}
%
%
%

The reason for adapting DGCNN, and not employing PointNet++ is to use the EdgeConv structure which has a small dense neighbor sampling. PointNet++ instead computes local PointNets and aggregates the features, where the point clouds vary in size from 512 to 128 \cite{qi2017pointnet++}, compared with ten to twenty in DGCNN \cite{dgcnn}. By using the EdgeConv we get a structure more similar to 2D convolutional networks with a dense sampling of points.
%
%
%
%


\subsubsection{Segmentation and offset regression}

Using the two final layers, both down-scaled to 2048 points, the object segmentation is computed. The segmentation is computed by processing each point with three consecutive MLPs. Along with the segmentation, the network also predicts the offset to the object center and the top point of the object. The center point is defined as the center of the object bounding box. The top point is determined by the axis of the bounding box along the symmetry axis. The structure of the developed network is shown in Fig.~\ref{fig:pyramid}.

\subsubsection{Input preprocessing}
\begin{figure}[t]
\centering
\subfloat[Point sampling.]{\includegraphics[width=.48\linewidth]{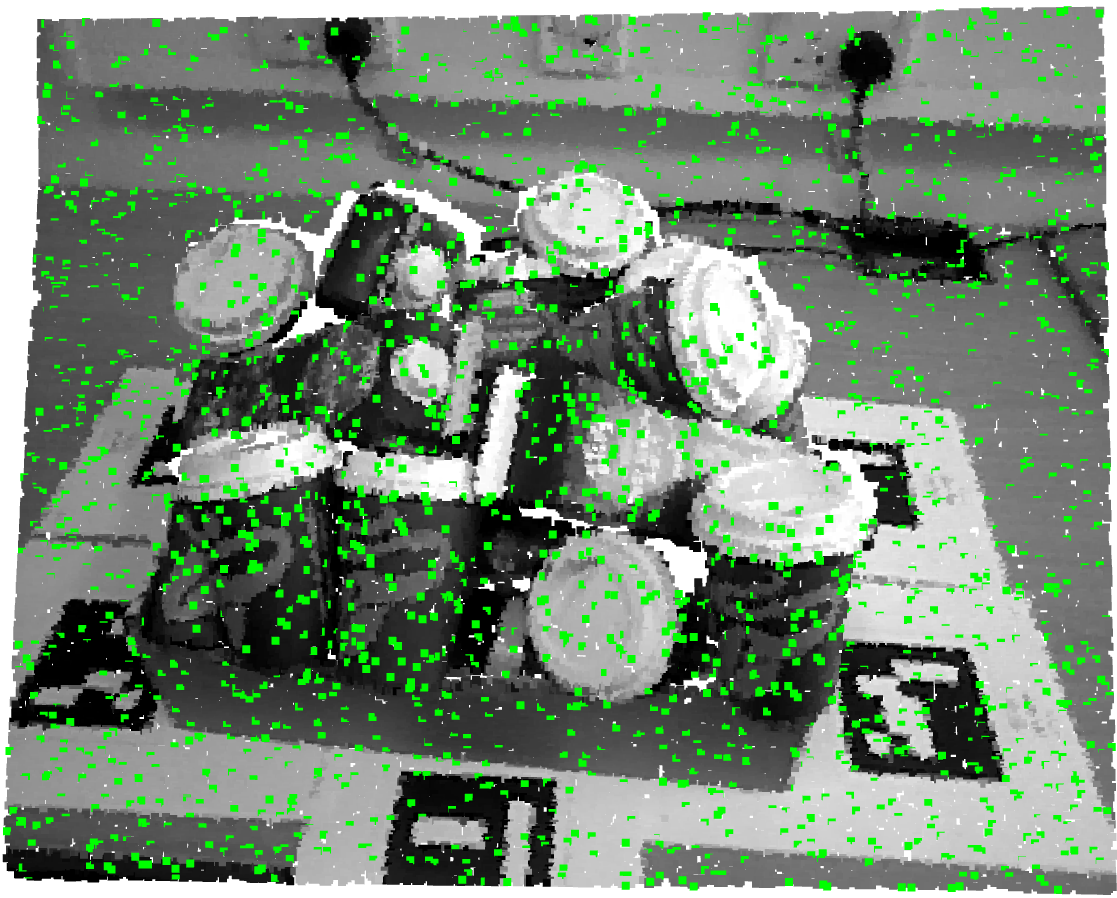}}\ 
\subfloat[Receptive field.]{\includegraphics[width=.48\linewidth]{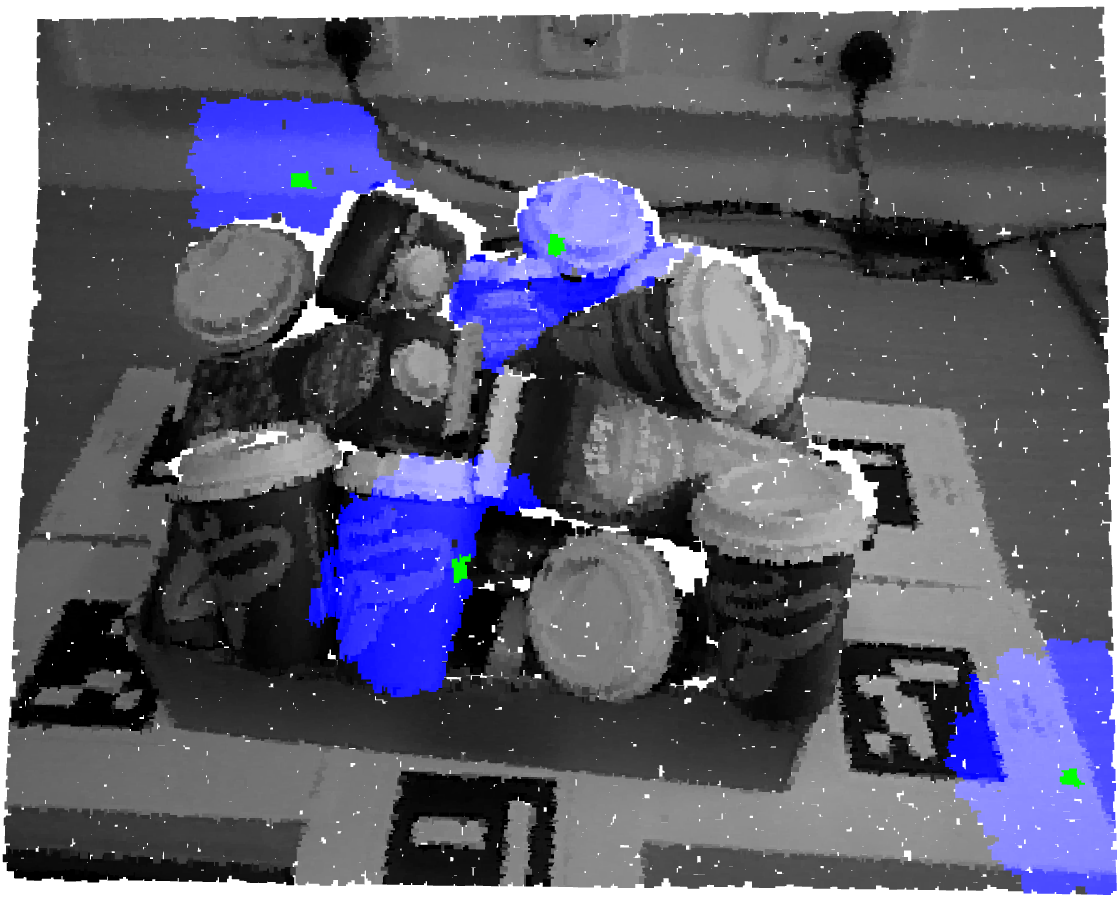}}
   \caption{Left (a): visualization of the density of the sampling  of the networks output points (green). Right (b): the receptive field (blue) for four output points (green).}
   \label{fig:sampling}
     \vspace{-6mm}
\end{figure}

The pre-processing receives a depth image and outputs the required information for the network, a point cloud with normal vectors, and neighbor indices for each layer. First, the depth image is converted to a point cloud. The point cloud is then down-scaled to 65,536 points, from which normal vectors are computed. The normal vectors are computed on the down-scaled point cloud to decrease the run-time. Neighbors are then computed for each down-scaled layer. The neighbors are computed by k-NN \cite{cover1967nearest} pooled from the previous layer. In the first layer, 16,384 points find neighbors from the 65,536 points. In the second layer, 8,192 points find neighbors from the 16,384 points, etc. The number of neighbors for each layer is set to twenty.
A difference to PointNet++ \cite{qi2017pointnet} is that instead of computing the down-sampling using Farthest Point Sampling (FPS), random down-sampling is used. While the random down-sampling does not guarantee the same evenly spread as FPS, it allows for much faster run-time. If FPS were used to sample from the large input point cloud several seconds would be required for processing severely limiting the usability of the method \cite{Li2022adjust}.
Our dense sampling along with the hierarchical down-sampling almost guarantees that all point information is still included. A visualization of the density of sampling and the receptive field are shown in Fig.~\ref{fig:sampling}.

\begin{figure}[t]
\centering
\subfloat[Center predictions.]{\includegraphics[width=.32\linewidth]{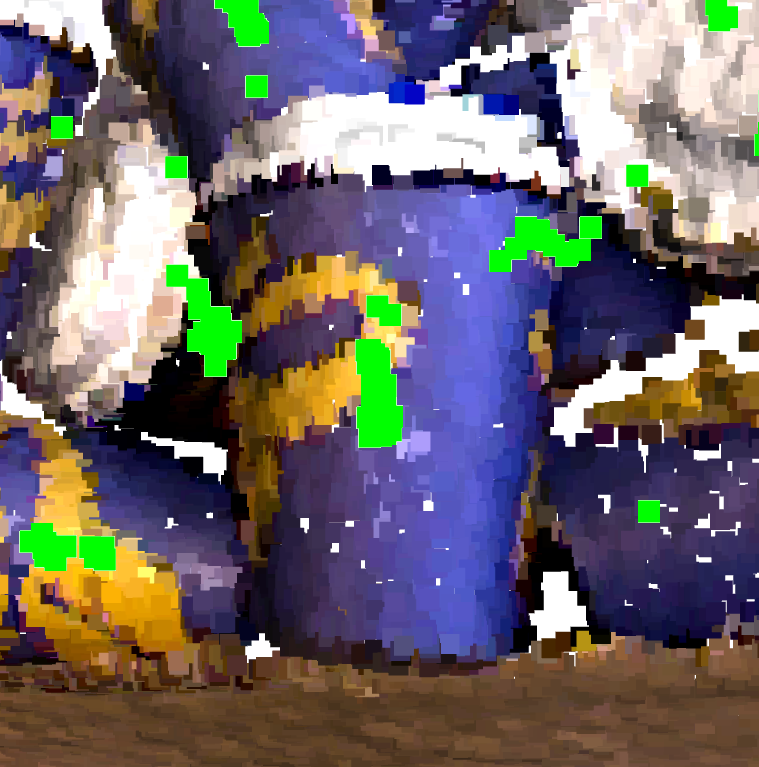}}\ 
\subfloat[Cluster center as object detection.]{\includegraphics[width=.32\linewidth]{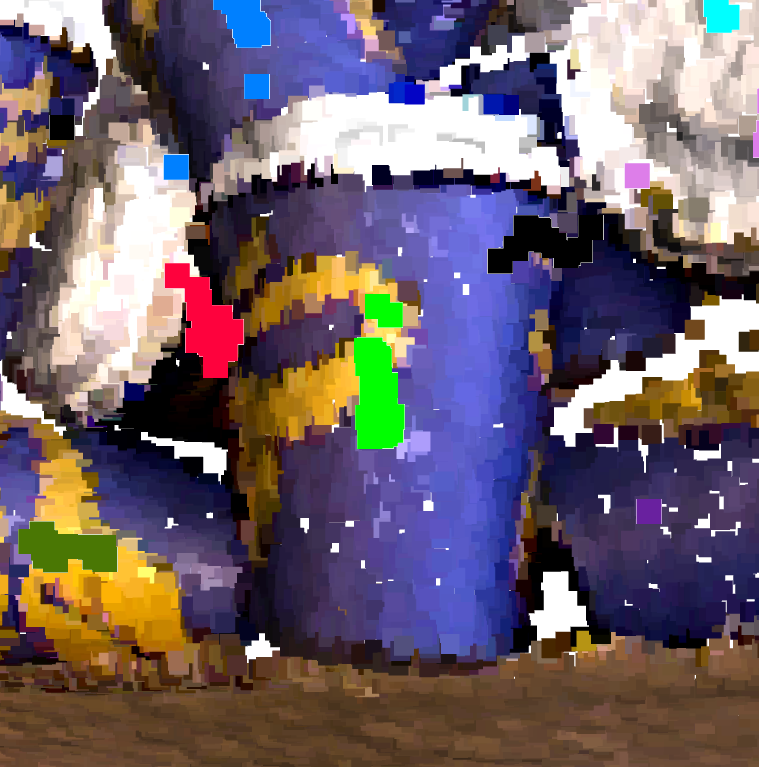}}\ 
\subfloat[Resulting instance segmentation.]{\includegraphics[width=.32\linewidth]{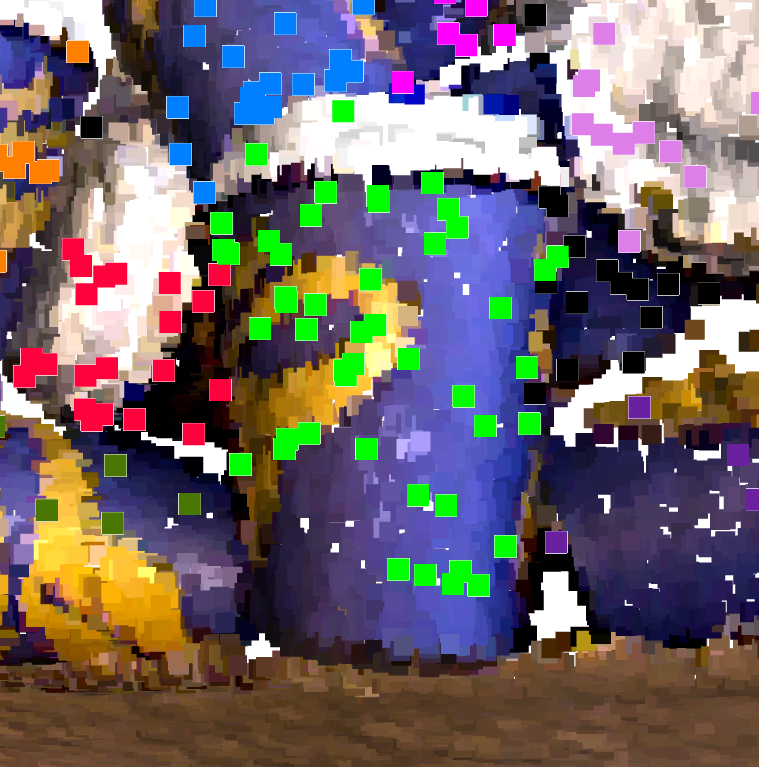}}

   \caption{Visualization of the object detection and segmentation process. (a) First, the center predictions are found by using the computed semantic segmentations and the center offsets. (b) The object detections are then computed by clustering the center predictions. (c) Finally, the instance segmentations are found by removing the center offset from the clustered points.}
   \label{fig:segvis}
     \vspace{-6mm}
\end{figure}



\subsubsection{Object detection from center predictions}
By combining the predicted center offsets with the segmentation results, predictions for object center points are found. For each predicted center point a non-maximum suppression (NMS) is performed. The NMS first computes all neighboring points within a threshold distance and the number of neighbors is calculated \cite{cover1967nearest}. The threshold distance is half the shortest side of the object bounding-box, ensuring that only points belonging to the same object instance are clustered together. Any point with a neighbor which has more neighbor points is removed. The remaining points from the NMS are the expected object centers, and all points within the threshold distance are part of this object. By removing the center offset this gives the points on the object surface. Resulting in both an instance segmentation and object detection. The object detection and resulting segmentations are shown in Fig.~\ref{fig:segvis}.

%
%


\subsubsection{5 DoF pose estimation from orientation vector}
The pose estimator uses both the center and top predictions. The top predictions are found similarly to the center prediction, by using the top-offsets.
NMS is performed on the top points to get the direction that most of the predictions agree on. The NMS is similar to the one employed for finding the center point. By subtracting the top point from the center point the orientation vector of the object is found, Eq.~(4). The full rotation matrix is then found by computing the x- and y-axis. The y-axis is found by computing the cross product between the orientation vector $\vec{n} $ and a tangent vector $\vec{tangent}$, Eq.~(5). The tangent vector is simply a vector which does not point in the direction of the orientation vector. Similarly, by computing the cross product of the orientation vector and the y-axis, the x-axis is found, Eq.~(6). By combining the x-axis, the y-axis, and the orientation vector the rotation matrix is created. The transform matrix is then finally found by combining the rotation matrix with the object center, Eq.~(7). 

\begin{align}
\vec{n} &= ||p_{center} - p_{top}|| \\
\vec{y} &= \vec{tangent} \times \vec{n} \\
\vec{x} &= \vec{y} \times \vec{n} \\
^{cam}T_{obj} &= \   
    \left[ {\begin{array}{cc}
   \{\vec{x},  \vec{y}, \vec{n}\} & p_{center} \\
   0 & 1 \\
  \end{array} } \right] 
\end{align}

\subsubsection{Segmentation-ICP}
To obtain a more precise pose estimation Iterative Closest Point (ICP) is employed \cite{arun1987least}. However, as the instance segmentation is available it is used for the refinement. Instead of matching the object point cloud with the full scene point cloud, only the segmented points are used. This allows the refinement to run much faster as fewer point distances needs to be calculated. Additionally, it will avoid refining to points not belonging to the object.
%
%

Another aspect is that the ICP process is reversed by fitting the scene point cloud to the object. Thus the reverse ICP finds the position of the segmented point cloud which best fits the object. By inverting the ICP we avoid matching parts of the object not visible in the scene, ensuring that the refinement fits with the entire segmentation.


\subsubsection{Extending to 6 DoF pose estimation}
The developed method only estimates a 5 DoF pose. To extend the method for 6 DoF pose estimation a z-axis pose refiner has been added. After the 5 DoF pose estimation the method rotates the object around the z-axis in intervals of $10\degree$ and performs the Segmentation-ICP. All refinements with less than $80\%$ overlap are discarded at this point. A depth projection \cite{hagelskjaer2021bridging} is then used to score each refinement. The depth projection measures overlap of the object projection with the real depth data, and is similar to the VSD \cite{hodan2018bop} score explained in Sec.~\ref{bop}.
We also test for wrong z-directions by rotating $180\degree$ about the x-axis. Lastly, the orientation search is limited by the pose range provided by the dataset.









\subsection{Network Training}
The network is trained on synthetic data provided for the BOP challenge \cite{hodavn2020bop}. In this dataset, the objects are dropped randomly along with other objects onto a surface. Camera views are then randomly generated and color images and depth images are created. The color images are generated using photorealistic image synthesis  \cite{hodavn2019photorealistic, hodavn2020bop}. However, the depth images are simply projections of the object's depth in the image. As our method only uses depth data it will overfit to the perfect synthetic data. 

To generalize the network to real data, data augmentation is applied to the depth image during training. Each of the different augmentations is applied with a preset random probability. The data augmentation consists of random circles cut out of the image, Gaussian noise, blurring, and random cropping of the image in x, y and z directions. Apart from the noise the cropping makes to the images, it also provides the network with different resolutions as the point cloud is then sampled from a smaller image. This provides the objects at different point densities during training.
This is opposed to PointNet++ where robustness to scale is obtained by different sampling strategies at test time \cite{qi2017pointnet++}. The last type of augmentation applied is the removal of edges. A Canny edge detector \cite{canny1986computational} is applied to the depth image, and the edges are used to remove depth information. This is to simulate real sensors where information is often lost close to the edges.

The network is trained for 100 iterations, using $95\%$ of the data for training and $5\%$ for validation. For the object segmentation cross-entropy is used to calculate the loss, where the center- and top-offsets simply use $L_1$ loss for the distance in mm. For the center- and top-offsets the loss is discarded if the point belongs to the background, and only calculated for object points. To help in the training a weighting of the loss is performed by multiplying the loss by $w_{type} \times 2 - w_{visibility}$, where $w_{visibility}$ is the visibility of the object according to the BOP dataset. This ensures that objects that are less visible are prioritized during training. Finally, $w_{type}$ is the Inverse Number of Samples strategy \cite{wang2017learning} for weighing each the loss of objects over the background as most points belong to the background.

\begin{figure}[t]
\centering
\subfloat[IC-BIN - Scene 1 - Image 18.]{\includegraphics[trim=0 0 0 0, clip, width=0.49\linewidth]{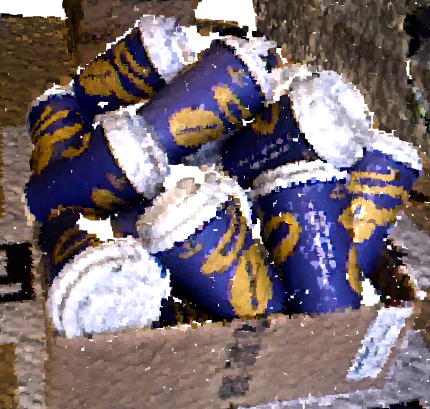} \includegraphics[trim=0 0 0 0,clip, width=0.49\linewidth]{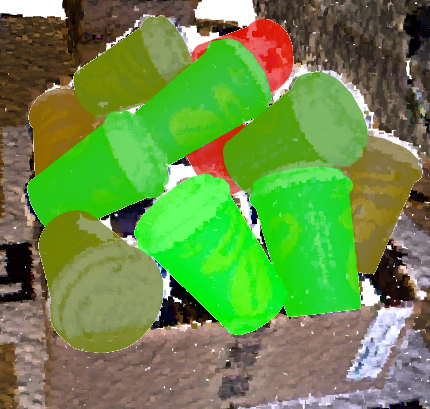}}

\subfloat[IC-BIN - Scene 2 - Image 28.]{\includegraphics[angle=-90, width=0.49\linewidth]{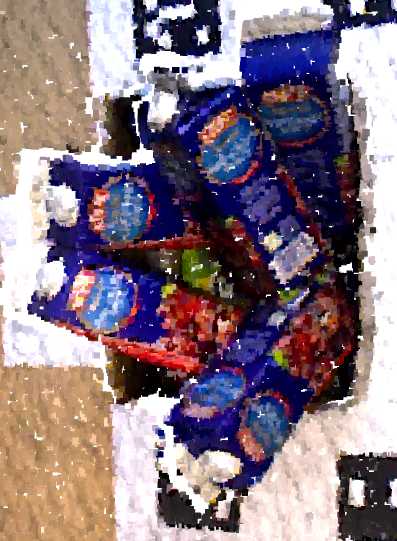} \includegraphics[angle=-90, width=0.49\linewidth]{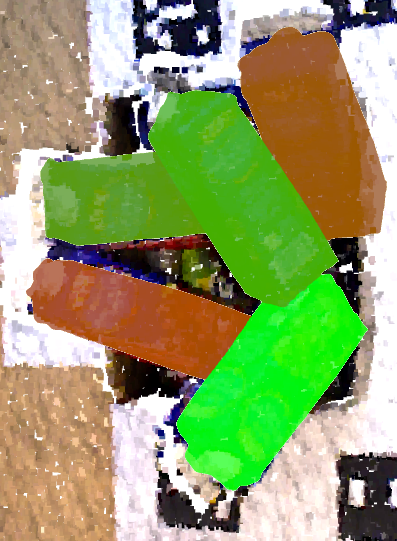}}

   \caption{Examples of pose estimation results on the test dataset. To the left the original image is shown and the pose estimations are shown to the right. The colors indicate the normalized amount of inliers for each pose estimation, from green to red.}
   \label{fig:peic}
     \vspace{-2mm}
\end{figure}


\section{Experiments}
To demonstrate the effectiveness of the developed method several experiments are performed. We show the performance on a benchmark dataset for bin picking pose estimation and compare it with several methods. We show the run-time contribution of each part of the algorithm and compare our Segmentation-ICP with classic ICP.

\subsection{IC-BIN}
Our developed method is tested on the IC-BIN dataset \cite{doumanoglou2016recovering}. This is a dataset with two different objects in three different scenes with approximately sixty images each. The objects are placed inside a bin with two homogeneous and one heterogeneous scene. The Coffee cup object is almost rotationally symmetric, and for this object, we employ the 5 DoF pose estimation, whereas the Juice object requires the 6 DoF method. The results of our method are shown in Tab.~\ref{tab:icbin}. It is seen that our method obtains state-of-the-art performance on the dataset and increases the average score from 57.6 to 71.8. Especially for the rotationally symmetric Coffee Cup, an increase of almost twenty percentage points is seen. Examples of pose estimations are shown in Fig.~\ref{fig:peic}.

\begin{table}[t]
\centering
   \caption{Pose estimation results on the IC-BIN dataset \cite{doumanoglou2016recovering}.}
   \label{tab:icbin}
\begin{tabular}{|c|c|c|c|}
\hline
    Method & Coffee & Juice & Average \\ \hline
    \cite{tejani2014latent} & 31.4 & 24.8  & 28.1 \\ \hline
    \cite{doumanoglou2016recovering} & 36.1   & 29.0  & 32.6 \\ \hline
    \cite{buch2017rotational} & 63.8   & 44.9  & 54.4 \\ \hline
    \cite{hagelskjaer2019bayesian} & 63.4   & 51.7  & 57.6 \\ \hline
    Ours & \textbf{83.6}  & \textbf{60.0}  & \textbf{71.8} \\ \hline
\end{tabular}
     \vspace{-2mm}
\end{table}

\subsubsection{Comparison on BOP}
\label{bop}
We also show performance on the IC-BIN dataset with the BOP metric \cite{hodavn2020bop, hodan2024bop}, while comparing with the state-of-the-art depth-only methods. 
We show the average score on the dataset and the Average Visual Surface Discrepancy, $AR_{VSD}$ \cite{hodan2018bop}. To calculate the VSD score the object is projected into the scene using the pose estimate and the ground truth pose. Using the depth image the visible part of the objects are calculated, and the depth distance is calculated between the estimate and ground truth. 
The VSD score is highlighted as this depth distance is invariant to symmetries in the object surface. 
The results are shown Tab.~\ref{tab:icbinbop}. From the results, it is seen that our method outperforms all other methods, especially when using only the $AR_{VSD}$ score.



\begin{table}[]
    \centering
   \caption{The overall score and VSD score for our method and the three best performing depth only methods on the IC-BIN BOP dataset \cite{hodan2024bop}. The first three comparison methods use PPF \cite{drost2010model}.}
   \label{tab:icbinbop}
    \begin{tabular}{|l|c|c|}
         \hline 
        Method               & $AR$ & $AR_{VSD}$ \\ \hline
        Drost-CVPR10-3D-Only & 0.388 & 0.501       \\ \hline
        Vidal-Sensors18    & 0.393 & 0.505       \\ \hline
        ZTE\_PPF          & 0.396 & 0.510       \\ \hline
        SMC \cite{redickbayesian}        & 0.316 & 	0.438       \\ \hline
        Ours                 & 	\textbf{0.417} & \textbf{0.561}      \\ \hline
    \end{tabular}
     \vspace{-3mm}
\end{table}

\begin{table}[t]
    \centering
    \caption{The run-time of a single pass of the algorithm using the Segmentation-ICP refinement.}
    \label{tab:cum_runtime}
    \begin{tabular}{|l|c|c|}
        \hline
        Task & Time (s) & Cum. Time (s) \\ \hline
        Loading point cloud &  0.013 & 0.013 \\
        Computing normals & 0.060 & 0.073 \\
        Computing neighbors & 0.128 & 0.201 \\
        Preparing for GPU & 0.002 & 0.203 \\
        Model processing & 0.005 & 0.208 \\
        Pose Estimation & 0.037 & 0.245 \\
        \hline
    \end{tabular}
         \vspace{-3mm}
\end{table}

\subsection{Run-time}
The focus of this paper is to develop a fast 5 DoF pose estimation method for full point clouds. To test the effectiveness of the developed method the run-time of each component in our pipeline has been recorded.  All experiments are performed on a PC environment (Intel Ultra 7 155H CPU and NVIDIA RTX 1000 Ada Generation Laptop GPU). The results for running a 5 DoF pose estimation with ten objects are shown in Tab.~\ref{tab:cum_runtime}. From Tab.~\ref{tab:cum_runtime} it is seen that a complete scene can be processed in less than 0.250 seconds. When looking at the individual components the greatest bottleneck is the neighbor computation and secondly the computation of normals. To improve the run-time of the algorithm these would be the most important focus. When running the 6 DoF pose estimation instead the run-time increases to 2.68 seconds, a very large increase in run-time. 

\subsection{Segmentation-ICP}
To test the effectiveness of the developed Segmentation-ICP, we compare it with classic ICP. The comparison is performed with the Coffee cup object, as pure ICP is only applied for this object. For both versions of ICP, the refinement is run until convergence. The run-time with the classic ICP increases from 0.037 to 0.365 seconds. And the score decreased from 83.6 to 79.6. By using our Segmentation-ICP the run-time decreased while the score increased.

\subsection{Number of neighbors}
The number of neighbors in the k-NN influences the receptive field of the network. To test the influence of the receptive field, we compare the performance when computing 10 and 40 neighbors. The results are shown in Tab.~\ref{tab:neighbors}. From the results, it is seen that performance decreases drastically with a decrease in receptive field. However, for 40 neighbors, the score increases for the larger Juice object, resulting in a greater overall score. But, the increased receptive field also increases the run-time of the neighbor computation from 0.128 to 0.202 seconds.

\begin{table}[t]
    \vspace{1.5mm}
\centering
   \caption{Pose estimation results on the IC-BIN dataset \cite{doumanoglou2016recovering} with varying number of neighbors.}
   \label{tab:neighbors}
\begin{tabular}{|c|c|c|c|}
\hline
    Neighbors & Coffee & Juice & Average \\ \hline
    10 & 65.8  & 39.2  & 52.5 \\ \hline
    20 & \textbf{83.6} & 60.0  & 71.8 \\ \hline
    40 & 82.6  & \textbf{65.3}  & \textbf{73.9} \\ \hline
\end{tabular}
     \vspace{-3mm}
\end{table}

\section{Conclusion}

In this paper, we have presented a network for the detection and pose estimation of cylindrical objects in colorless point clouds. We believe this method has many applications in industry. Our method was tested on the IC-BIN benchmark dataset and demonstrated state-of-the-art performance.  In further work, the method will be implemented on a robotic set-up and used for pose estimation in a bin-picking scenario. The algorithm could also be extended to 6 DoF pose estimation. However, instead of predicting an orthogonal vector, the uncertainty of each possible rotation can be determined \cite{murphy2021implicit}. This would allow the model to incorporate the inherent aleatoric ambiguities and provide a better estimation of the poses.

\bibliographystyle{IEEEtran} 
\bibliography{egbib}

\end{document}